%% file: root.tex
\documentclass[sigconf, noacm]{acmart}

\AtBeginDocument{%
  \providecommand\BibTeX{{%
    \normalfont B\kern-0.5em{\scshape i\kern-0.25em b}\kern-0.8em\TeX}}}

\newtheorem{theorem}{Theorem}[section]

\newtheorem{problem}{Problem}
\newtheorem{definition}[theorem]{Definition}
\newtheorem{remark}[theorem]{Remark}

\usepackage{graphicx} 
\usepackage{epsfig} 
\usepackage{amsmath}
\usepackage{epstopdf}
\usepackage{multirow}
\usepackage{rotating}
\usepackage{mathrsfs}
\usepackage{color}
\usepackage{bbm}
\usepackage{enumerate}
\usepackage{fancyhdr}
\usepackage{algorithm}
\usepackage{algorithmic}
\usepackage{graphicx}
\usepackage{listings}
\usepackage[]{footmisc}
\usepackage{textcomp}
\usepackage{caption}
\usepackage{subcaption}
\usepackage{ulem}
\usepackage{booktabs}

\usepackage{tikz}
\usetikzlibrary{decorations.markings}

\usepackage{todonotes}
\newcommand{\rom}[1]{\uppercase\expandafter{\romannumeral #1\relax}}

\makeatletter
\def\BState{\State\hskip-\ALG@thistlm}
\makeatother

\DeclareMathOperator*{\argmax}{argmax}

\renewcommand\footnotetextcopyrightpermission[1]{}

\begin{document}

\title{
Offline Learning of Closed-Loop Deep Brain Stimulation Controllers for Parkinson Disease Treatment
}

\author{Qitong Gao}
\affiliation{
 \department{Electrical and Computer Engineering}
 \institution{Duke University}
  \city{Durham}
  \state{NC}$  $
  \postcode{27708}
  \country{USA}
}
\email{qitong.gao@duke.edu}

\author{Stephen L. Schmidt}
\affiliation{
 \department{Biomedical Engineering}
 \institution{Duke University}
  \city{Durham}
  \state{NC}$  $
  \postcode{27708}
  \country{USA}
}
\email{stephen.schmidt@duke.edu}

\author{Afsana Chowdhury}
\affiliation{%
 \department{Electrical and Computer Engineering}
 \institution{Duke University}
  \city{Durham}
  \state{NC}$  $
  \postcode{27708}
  \country{USA}
}
\email{afsana.chowdhury@duke.edu}

\author{Guangyu Feng}
\affiliation{%
 \department{Electrical and Computer Engineering}
 \institution{Duke University}
  \city{Durham}
  \state{NC}$  $
  \postcode{27708}
  \country{USA}
}
\email{guangyu.feng@duke.edu}

\author{Jennifer J. Peters}
\affiliation{%
 \department{Biomedical Engineering}
 \institution{Duke University}
  \city{Durham}
  \state{NC}$  $
  \postcode{27708}
  \country{USA}
}
\email{jennifer.peters@duke.edu}

\author{Katherine Genty}
\affiliation{%
 \department{Neurosurgery}
 \institution{Duke University}
  \city{Durham}
  \state{NC}$  $
  \postcode{27708}
  \country{USA}
}
\email{katherine.genty@duke.edu}

\author{Warren M. Grill}
\affiliation{%
 \department{Biomedical Engineering}
 \institution{Duke University}
  \city{Durham}
  \state{NC}$  $
  \postcode{27708}
  \country{USA}
}
\email{warren.grill@duke.edu}

\author{Dennis A. Turner}
\affiliation{%
 \department{Neurosurgery}
 \institution{Duke University}
  \city{Durham}
  \state{NC}$  $
  \postcode{27708}
 \country{USA}
}
\email{dennis.turner@duke.edu}

\author{Miroslav Pajic}
\affiliation{%
 \department{Electrical and Computer Engineering}
 \institution{Duke University}
  \city{Durham}
  \state{NC}$  $
  \postcode{27708}
  \country{USA}
}
\email{miroslav.pajic@duke.edu}
\thanks{This work is sponsored in part by the
NSF CNS-1837499 award and the National AI Institute for Edge Computing Leveraging Next Generation Wireless Networks, Grant CNS-2112562, as well as by NIH UH3 NS103468. Investigational Summit RC+S systems and technical support provided by Medtronic PLC. Apple Watches were provided by Rune Labs.}

\renewcommand{\shortauthors}{Qitong Gao et al.}




\copyrightyear{2023} 
\acmYear{2023} 
\acmConference[ICCPS '23]{14th ACM/IEEE International Conference on Cyber-Physical Systems (with CPS-IoT Week 2023)}{May 9--12, 2023}{San Antonio, Texas}
\acmBooktitle{14th ACM/IEEE International Conference on Cyber-Physical Systems (with CPS-IoT Week 2023) (ICCPS '23), May 9--12, 2023, San Antonio, Texas}

\begin{abstract}
Deep brain stimulation (DBS) has shown great promise toward treating motor symptoms caused by Parkinson's disease (PD), by delivering electrical pulses to the Basal Ganglia (BG) region of the brain. However, 
DBS devices approved by the U.S. Food and Drug Administration (FDA) can only deliver continuous DBS (cDBS) stimuli at a fixed amplitude; this
energy inefficient operation reduces battery lifetime of the device, 
cannot adapt treatment dynamically for activity,
and may cause 
significant side-effects (\textit{e.g.,} gait impairment). 
In this work, we introduce an offline reinforcement learning (RL) framework, allowing the use of past clinical data to train an RL 
policy to adjust the stimulation amplitude in real time, with the goal of reducing energy use while maintaining the same level of treatment (\textit{i.e.}, control) efficacy as cDBS. Moreover, clinical protocols require the safety and performance of such RL controllers to be demonstrated ahead of deployments in patients. Thus, we also introduce an offline policy evaluation (OPE) method 
to estimate the performance of RL policies using historical data, before deploying them on patients. 
We evaluated our framework on four PD patients equipped with the 
RC+S DBS system, 
employing the RL controllers 
during monthly clinical visits, with the overall \textit{control efficacy} evaluated by severity of symptoms (\textit{i.e.}, bradykinesia and tremor), changes in PD biomakers (\textit{i.e.}, local field potentials), and patient ratings. 
The results from clinical experiments show that our RL-based controller 
maintains the same level of control efficacy as cDBS, but with significantly reduced stimulation energy. 
Further, the OPE method is shown effective in accurately estimating and ranking the expected returns of 
RL controllers. 
\end{abstract}

\maketitle


\input{Intro.tex}

\input{Preliminary.tex}

\input{RL.tex}

\input{OPE.tex}

\input{Evaluation.tex}
\input{Discussion.tex}

\vspace{-4pt}
\bibliographystyle{ACM-Reference-Format2}
\bibliography{dbs_references}

\appendix

\section{Notation}
Here, we define the notation 
used in the paper. The sets of reals, integers, and positive integers, are denoted by $\mathbb{R}$, $\mathbb{Z}$, and $\mathbb{Z}^+$, respectively. Further, $x \sim p(x)$ denotes that random variable $x$ is sampled from distribution $p(x)$. We also use $\mathcal{N}(x;\mu,\mathbf{\Sigma})$ to denote Gaussian distributions with mean $\mu$ and covariance matrix $\mathbf{\Sigma}$ over variable $x$. For simplicity, we write $x \sim \mathcal{N}(\mu,\mathbf{\Sigma})$ during sampling. The KL-divergence between distributions $p(x)$ and $q(x)$ is defined~as
    \begin{align}
        KL(p||q) = \mathbb{E}_p\left[\log\frac{q(x)}{p(x)}\right].
    \end{align}

{
\section{Availability of Data and Code}
We plan to open-source the data collected from clinical testing, as well as the implementation for training RL policies, in the future\footnote{After finalizing a journal submission built on top of this work.}, to facilitate research in developing RL-based DBS controllers. The RC+S system as well as its Summit code base are considered proprietary, which may not be published online. The implementation of our OPE method, DLSM, is built on our previous works~\cite{gao2022offline, gao2023variational}, with code published at \texttt{\url{https://github.com/gaoqitong/vlbm}}.

\section{Additional Preliminaries}
Below we introduce in details the preliminaries needed to supplement Sec.~\ref{sec:prelim}.
\subsection{Deep Actor-Critic RL}
\label{app:deep_ac}
We now briefly introduce the deep actor-critic algorithm~\cite{lillicrap2015continuous} and refer the readers to~\cite{gao2020model, gao2022offline, lillicrap2015continuous} for more details. 
First, the state-action value functions can be defined as follows.

\begin{definition}[State-Action Value Function]\label{def:q_function}
Given~an MDP $\mathcal{M}$ and policy $\pi$, the state-action value function $Q^\pi (s , a )$, where $s \in \mathcal{S}$ and $a \in \mathcal{A}$, is defined as the expected return for taking action $a $ when at state $s $ following policy $\pi$ at stage $t$, i.e., \begin{equation}
Q^\pi (s , a ) = \mathbb{E}_{s\sim \mathcal{S}, a \sim \mathcal{A}}[G_t|s _t = s , a_t = a ].
\end{equation}
\end{definition}

Two neural networks, with weights $\theta_a$ and $\theta_c$, can be used to parameterize the policy (actor) $\pi_{\theta_a}(s):\mathcal{S} \rightarrow \mathcal{A}$ and the Q-functions (critic) $Q_{\theta_c}(s,a):\mathcal{S} \times \mathcal{A} \rightarrow \mathbb{R}$, respectively. Finally, the target policy $\pi^*=\pi_{\theta_a^*}$ can be obtained by optimizing over
\begin{align}
\label{eq:ac_objective}
    \max_{\theta_a, \theta_c} \mathbb{E}_{s,a,r,s'\sim\mathcal{E}^\mu}\left[Q_{\theta_c}\left(s,\pi_{\theta_a}(s)\right)\right];
\end{align}
this can be achieved using gradient descent, over all the training samples in the experience replay buffer $\mathcal{E}^\mu$~\cite{lillicrap2015continuous}.

\subsection{Deep Latent MDP Model (DMLL)}
\label{app:dmll}

The DLMM is trained to fit the transitions of the MDP $p(s_{t+1}|s_t,a_t)$ and rewards $r_t = R(s_t, a_t)$, which consist of three components, \textit{i.e.}, a latent prior, posterior and sampling distribution. The prior $p_\psi(z_t)$ is parameterized by $\psi$, over the latent variable space (LVS) $\mathcal{Z} \subset \mathbb{R}^d$, where $d \in \mathbb{Z}^+$ is the dimension. The prior represents one's belief over the latent distribution of the states (\textit{i.e.}, probability density function over the latent variables) which is considered unknown; thus, it is usually chosen to be a multivariate Gaussian with zero mean and identity covariance. Then, the encoder (or approximated posterior) $q_\phi(z_t|s_t)$ is parameterized by $\phi$, which is responsible for encoding the MDP state $s_t \in \mathcal{S}$ into the LVS, $\mathcal{Z}$. Note that the true posterior $p_\psi(z_t|s_t)$ is intractable, since its density function contains integration over $\mathcal{Z}$ which is deemed unknown; we refer 
to~\cite{gao2022offline} for more details. Lastly, the decoder (or sampling distribution) $p_\psi(s_{t+1}, r_{t}|$ $z_t, a_t)$ enforces the MDP transition from $t$~to~$t+1$. 

Hence, the DLMM can be used to interact with the policy $\pi$ and generate simulated trajectories via 
\vspace{-4pt}
\begin{align}
    z_t \sim q_\phi(z_t|\hat s_t), \quad
    \hat s_{t+1}, \hat  r_{t} \sim p_\psi(s_{t+1},r_{t}|z_t,a_t); \label{eq:dlmdp_sampling_1}
\end{align}
here, $\hat s_t, \hat s_{t+1}$ and $\hat r_t$ represent the states and rewards predicted from DLMM. Consequently, the expected return of $\pi$ can be estimated as $\frac{1}{M}\sum\nolimits_{i=1}^M \sum\nolimits_{t=0}^T \gamma^t \hat r_t^{(i)}$, where $M$ is the total number of simulated trajectories generated following the process above, and $\hat r_t^{(i)}$ is the predicted reward at step $t$ in the $i$-th simulated trajectory. To train DLMM, one can maximize the evidence lower bound (ELBO) of the joint log-likelihood $\sum\nolimits_{t=0}^T \log p_\psi(s_{t+1},r_t)$, where the derivation of the ELBO for DLMM can be found in~\cite{gao2022offline}.

From~\eqref{eq:dlmdp_sampling_1}, it can be observed that the predicted $\hat s_{t+1}, \hat r_t$ are conditioned on $z_t$'s, which are dependent on the predicted state $\hat s_t$ from the last step. As a result, such an iterative process may not scale well to environments with longer horizons and more complicated dynamics, as the prediction error from all earlier steps are propagated into the future steps. Moreover, this DLMM is not capable of predicting the QoC metrics that are only evaluated \textit{once at the end of each session}, including the bradykinesia results, patient ratings and tremor severity as discussed in Sec.~\ref{subsec:setup}. To address such limitations, we introduce a new latent modeling method in Sec.~\ref{sec:OPE}, which decouples the dependencies between $z_t$'s and $\hat s_t$'s by directly enforcing the temporal transitions over latent variables,~\textit{i.e.}, $p(z_{t+1}|z_t, a_t)$.

}

\begin{algorithm}[!t]
\small
\caption{Train DLSM.}\label{alg}
\begin{algorithmic}[1]
\REQUIRE Model weights $\psi, \phi$, experience replay buffer $\mathcal{E}^\mu$, and learning rate $\alpha$.
\ENSURE
\STATE Initialize $\psi, \phi$
\FOR {$iter$ in $1:max\_iter$}
\STATE Sample a trajectory $[(s_0,a_0,r_0,s_1),\dots,$ $(s_{T-1},a_{T-1},r_{T-1},s_T)]\sim\mathcal{E}^\mu$
\STATE $z_0^\phi \sim q_\phi(z_0|s_0)$
\STATE $z_0^{\psi} \sim p_\psi(z_0)$
\STATE Run forward pass of DLSM following~(\ref{eq:encoder}) and~(\ref{eq:decoder}) for $t=1:T$, and collect all variables needed to evaluate the all terms within the expectation in $\mathcal{L}_{ELBO}$, which is denoted as $\tilde{\mathcal{L}}_{ELBO}$.
\STATE $\psi \gets \psi + \alpha \nabla_\psi \tilde{\mathcal{L}}_{ELBO}$
\STATE $\phi \gets \phi + \alpha  \nabla_{\phi} \tilde{\mathcal{L}}_{ELBO}$
\ENDFOR
\end{algorithmic}
\vspace{-2pt}
\end{algorithm}

\section{Algorithm to Train DLSM}
\label{app:alg}

Here we introduce how to use gradient descent to maximize the ELBO~\eqref{eq:elbo}, resulting in Algorithm~\ref{alg}.
For simplicity, we first illustrate with the case where the training batch only contains a single trajectory, and then extend to the cases where each batch contain $n$ trajectories. In each iteration, a trajectory is sampled from the experience replay buffer $\mathcal{E}^\mu$. Then, the initial latent state in the encoder is obtained following $z_0^\phi \sim q_\phi(z_0|s_0)$, while the initial latent state for the sampling distribution is generated following the latent prior $z_0^{\psi} \sim p_\psi(z_0)$. The processes introduced in~\eqref{eq:encoder} can be used to generate $z_t^\phi$'s iteratively given $z_0^\phi$. Similarly, $s_t^\psi, r_t^\psi, z_t^\psi$ can be generated iteratively following~\eqref{eq:decoder}. As a result, the log-likelihoods and KL-divergence terms within the expectation in $\mathcal{L}_{ELBO}$, defined in~\eqref{eq:elbo}, can be evaluated using the variables above, after which $\psi,\phi$ can be updated using the gradients $\nabla_\psi\tilde{\mathcal{L}}_{ELBO},\nabla_\phi\tilde{\mathcal{L}}_{ELBO}$, respectively, where $\tilde{\mathcal{L}}_{ELBO}$ refers to all the terms within the expectation in $\mathcal{L}_{ELBO}$. This algorithm is summarized in Alg.~\ref{alg}.

To extend to batch gradient descent, in line~3 of Algorithm~\ref{alg}, a batch of $n$ trajectories, $\mathcal{B}(n)$, will be sampled,~\textit{i.e.},
\begin{align}
    \mathcal{B}(n) = \Big[&\big[(s_0^{(1)},a_0^{(1)},r_0^{(1)},s_1^{(1)}),\dots,(s_{T-1}^{(1)},a_{T-1}^{(1)},r_{T-1}^{(1)},s_T^{(1)})\big],\dots,\nonumber\\
    &\big[(s_0^{(n)},a_0^{(n)},r_0^{(n)},s_1^{(n)}),\dots,(s_{T-1}^{(n)},a_{T-1}^{(n)},r_{T-1}^{(n)},s_T^{(n)})\big]\Big]\sim\mathcal{E}^\mu .
\end{align}
Then, the processes illustrated in lines 4-6 in Algorithm~\ref{alg} can be executed in $n$ parallel threads, with each corresponding to a unique trajectory in $\mathcal{B}(n)$. Further, then, $\tilde{\mathcal{L}}_{ELBO}$ can be evaluated as
\begin{align}
    \tilde{\mathcal{L}}&_{ELBO} (\psi,\phi) =  \frac{1}{n} \sum\nolimits_{i=1}^n \Big[\sum\nolimits_{t=0}^T \log p_\psi(s_t^{(i)}|z_t^{(i)}) \nonumber\\ & + \sum\nolimits_{t=1}^T \log p_\psi(r_{t-1}^{(i)}|z_t^{(i)}) \nonumber\\ & + \log p_\psi(r_{end}^{(i)}|z_T^{(i)}) - KL\big(q_\phi(z_0^{(i)}|s_0^{(i)}) || p(z_0^{(i)})\big)  \nonumber\\
    & - \sum\nolimits_{t=1}^T  KL\big(q_\phi(z_t^{(i)}|z_{t-1}^{(i)},a_{t-1}^{(i)},s_t^{(i)})||p_\psi(z_t^{(i)}|z_{t-1}^{(i)},a_{t-1}^{(i)})\big)  \Big],
\end{align}
with $s_t^{(i)}, a_t^{(i)}, z_t^{(i)}, r_t^{(i)}, r_{end}^{(i)}$ being the variables involved in one of the threads above processing the $i$-th trajectory in $\mathcal{B}(n)$.

\section{Participant Characteristics}
\label{app:patient}

\paragraph{Participant 1}
Episodes of tremor only. Bradykinesia in the left hand correlates with right STN beta amplitude. STN beta amplitudes in both hemispheres correlate with stimulation amplitude. {This participant takes 2 tablets of Sinemet 25mg/100mg for each clinical visit, one at 2 hrs before the testing begins (~7am) and another in the middle of the visit (around noon) respectively.}

\paragraph{Participant 2}
Very large amplitude tremor returns within seconds of low amplitude DBS. Bradykinesia in right hand highly correlated with left STN beta amplitude. Left STN beta amplitude also correlated with DBS amplitude. {This participant takes Flexeril 10mg, Selegiline 5mg and Pramipexole 0.125mg (1 tablet for each) at 2 hrs before the testing begins.}

\paragraph{Participant 3}
Pronounced tremor (hands and jaw) returns within seconds of low amplitude closed-loop DBS. Bradykinesia in right hand correlates with left STN beta amplitude. Left STN beta amplitude is the most responsive to stimulation of the cohort. {This participant takes 2 tablets of Sinemet 25mg/100mg for each clinical visit, one at 2 hrs before the testing begins (~7am) and another in the middle of the visit (around noon) respectively.}

\paragraph{Participant 4}
Pronounced tremor. Monopolar left STN DBS can produce a dyskinesia in neck, so clinical settings have been a bipolar configuration. Right hand bradykinesia correlated to left STN beta~power. {This participant takes 1 tablets of Sinemet 25mg/100mg at 2 hrs before the testing begins (~7am).}

\section{Proof of Theorem~\ref{thm}}
\label{app:proof}
We now derive the evidence lower bound (ELBO) for the joint log-likelihood distribution, \textit{i.e.},
\begin{align}
& \log p_\psi (s_{0:T},r_{0:T-1},r_{end}) \\ 
= & \log \int_{z_{1:T} \in \mathcal{Z}} p_\psi (s_{0:T},z_{1:T},r_{0:T-1},r_{end}) dz \\ 
 = & \log \int_{z_{1:T} \in \mathcal{Z}} \frac{p_\psi (s_{0:T},z_{1:T},r_{0:T-1},r_{end})}{q_\phi(z_{0:T}|s_{0:T}, a_{0:T-1})} q_\phi(z_{0:T}|s_{0:T}, a_{0:T-1}) dz \label{eq:derive_elbo_1}\\
 \geq & \mathbb{E}_{z_t\sim q_\phi} [\log p(z_0) + \log p_\psi (s_{0:T},z_{1:T},r_{0:T-1}|z_0) + \log p_\psi(r_{end}|z_T) \nonumber\\ & \quad\quad\quad - \log q_\phi(z_{0:T}|s_{0:T}, a_{0:T-1})] \label{eq:derive_elbo_2}\\
 = & \mathbb{E}_{z_t\sim q_\phi} \Big[\log p(z_0) + \log p_\psi(s_0|z_0) + \log p_\psi(r_{end}|z_T) \nonumber\\ & \quad\quad\quad + \sum\nolimits_{t=1}^T \log p_\psi(s_t,z_t,r_{t-1}|z_{t-1},a_{t-1}) \nonumber\\ & \quad\quad\quad - \log q_\phi(z_0|s_0) - \sum\nolimits_{t=1}^T \log q_\phi(z_t|z_{t-1},a_{t-1},s_{t}) \Big]\\
 = & \mathbb{E}_{z_t\sim q_\phi} \Big[\log p(z_0) - \log q_\phi(z_0|s_0) + \log p_\psi(s_0|z_0) + \log p_\psi(r_{end}|z_T) \nonumber\\ & \quad\quad\quad + \sum\nolimits_{t=1}^T \log \big(p_\psi(s_t|z_t)p_\psi(r_{t-1}|z_t)p_\psi(z_t|z_{t-1},a_{t-1})\big) \nonumber\\ & \quad\quad\quad  - \sum\nolimits_{t=1}^T \log q_\phi(z_t|z_{t-1},a_{t-1},s_{t}) \Big] \\
  = & \mathbb{E}_{z_t\sim q_\phi} \Big[\sum\nolimits_{t=0}^T \log p_\psi(s_t|z_t) + \log p_\psi(r_{end}|z_T) \nonumber\\ & \quad\quad\quad + \sum\nolimits_{t=1}^T \log p_\psi(r_{t-1}|z_t) -KL\big(q_\phi(z_0|s_0) || p(z_0)\big) \nonumber\\ 
  & \quad\quad\quad   - \sum\nolimits_{t=1}^T KL\big(q_\phi(z_t|z_{t-1},a_{t-1},s_{t})||p_\psi(z_t|z_{t-1},a_{t-1})\big)\Big].
\end{align}
Note that the transition from~\eqref{eq:derive_elbo_1} to~\eqref{eq:derive_elbo_2} follows Jensen's inequality.



\section{OPE Metrics}
\label{app:ope_metrics}

\paragraph{Rank correlation}
Rank correlation measures the Spearman's rank correlation coefficient between the ordinal rankings of the estimated returns and actual returns across policies,~\textit{i.e.},\\
$
    \rho = \frac{Cov(\mbox{rank}(V^{\pi}_{1:P}), \mbox{rank}(\hat{V}^{\pi}_{1:P}))}{\sigma(\mbox{rank}(V^{\pi}_{1:P})) \sigma(\mbox{rank}(\hat{V}^{\pi}_{1:P}))},
$
where $\mbox{rank}(V^{\pi}_{1:P})$ is the ordinal rankings of the actual returns, and $\mbox{rank}(\hat{V}^{\pi}_{1:P})$ is the ordinal rankings of the OPE-estimated returns.

\paragraph{Regret@1}
Regret@1 is the (normalized) difference between value of the actual best policy, against value of the policy associated with the best OPE-estimated return, which is defined as
$(\max_{i\in1:P} V^{\pi}_i-\max_{j\in\mbox{best}(1:P)} V^{\pi}_j)/\max_{i\in1:P} V^{\pi}_i$
where $\mbox{best}(1:P)$ denotes the index of the best policy over the set of $P$ policies as measured by estimated values $\hat V^{\pi}$.

\paragraph{Mean Absolute error (MAE)}
MAE is defined as the absolute difference between the actual return and estimated return of a policy:
$MAE = |V^{\pi}-\hat{V}^{\pi}|;$
here, $V^{\pi}$ is the actual value of the policy $\pi$, and $\hat{V}^{\pi}$ is the estimated value of $\pi$.

\end{document}

%% file: Intro.tex


\section{Introduction}
\label{sec:intro}

\begin{figure}[!t]
    \centering
    \includegraphics[width=0.74\columnwidth]{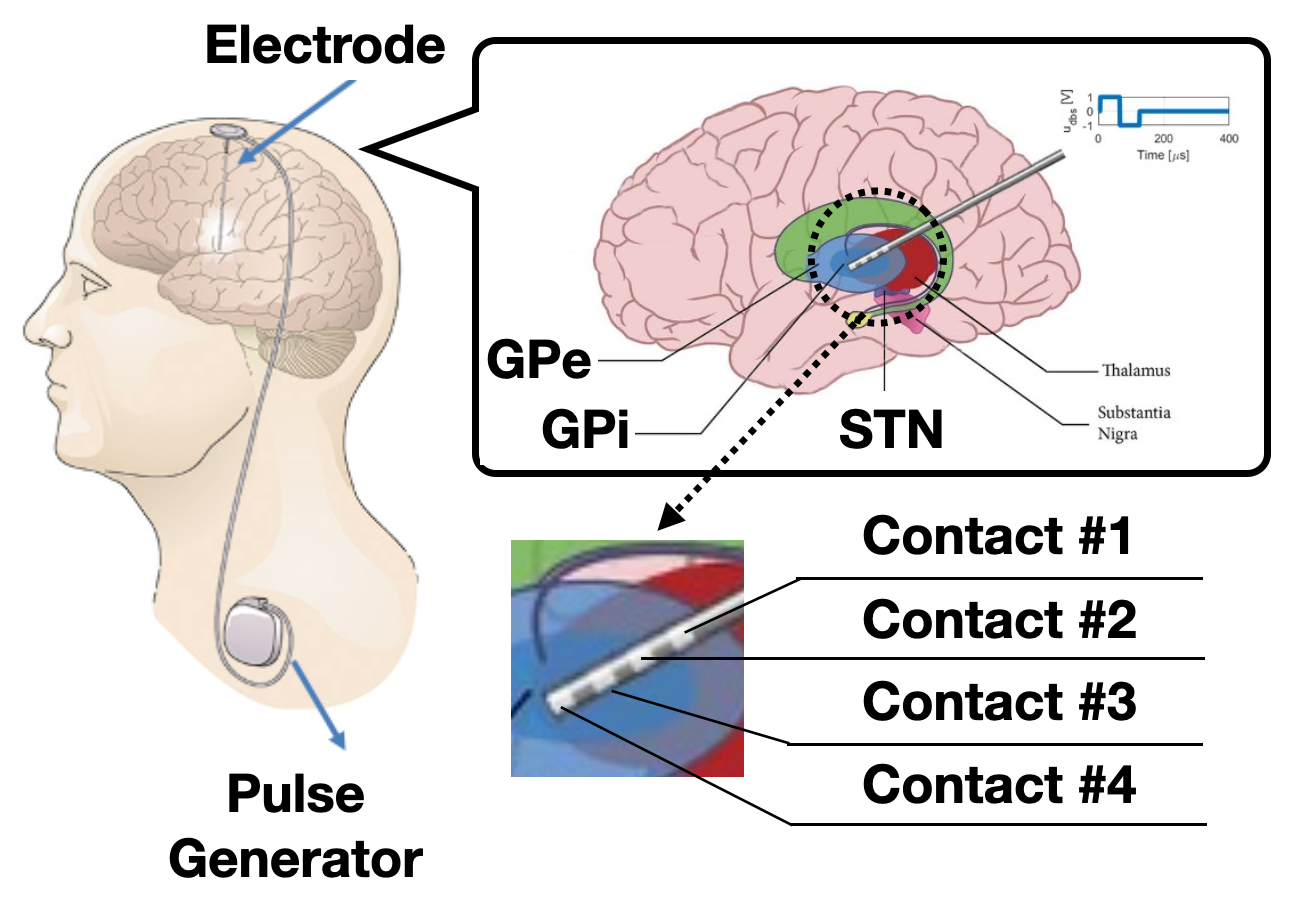}
    \vspace{-10pt}
    \caption{An implantable deep brain stimulation (DBS) device. The 
    stimuli, generated by the pulse generator at a given amplitude and frequency, are delivered to the basal ganglia (BG) through multi-contact electrodes. Each electrode 
    has four contacts; two  stimulate the BG and two sense local field potentials (LFPs) 
    that may be used 
    for control feedback.}
    \label{fig:Regions1mod}
\end{figure}

Currently, around 1.05 million individuals in the United States are affected by Parkinson's disease (PD)~\cite{marras2018prevalence}. Deep brain stimulation (DBS) is an effective treatment 
to reduce PD symptoms such as tremor and bradykinesia~\cite{benabid2003deep, deuschl2006randomized, follett2010pallidal, okun2012deep}. 
A DBS system consists of electrodes that are placed into the Basal Ganglia (BG) region of the brain, and a pulse generator implanted in the chest 
to generate trains of short electrical pulses (see Fig.~\ref{fig:Regions1mod}). Existing FDA-approved DBS solutions are limited to continuous DBS (cDBS). These devices are programmed to stimulate at a fixed amplitude, with the specific parameters determined by clinicians through trial-and-error~\cite{pineau2009treating}. However, such stimuli usually lead to extensive energy consumption, 
significantly reducing the battery lifetime of the device. Moreover, 
over-stimulated patients, even intermittently, may suffer from side-effects such as dyskinesia and speech impairment~\cite{beudel2016adaptive}. As a result, developments of closed-loop DBS controllers that are more responsive to activity and patient state (i.e., context) are of considerable interest to clinicians, patients, and the community.

Existing DBS control methods focus on simply switching on/off the stimulation or scaling up/down its intensity in a proportional control approach, conditioned on the change of specific biomarkers, \textit{i.e.}, when they cross over some pre-determined thresholds~\cite{beudel2016adaptive, arlotti2016external, arlotti2016adaptive, little2016adaptive, little2013adaptive}. Biomarkers include local field potentials (LFPs) and electroencephalography (EEG) from the~BG, as well as accelerometery data 
and electromyography 
obtained from wearable devices~\cite{opri2020chronic}. Though such methods have improved energy efficiency~\cite{habets2018update, little2016adaptive}, they still require substantial efforts 
to experiment and fine-tune the thresholds for each specific patient. 
Moreover, the patient may suffer from sub-optimal DBS settings in between clinical visits with poor symptom control due to varying patient state. For example, exercise or fluctuations in medication dosage or timing could affect their PD symptoms and DBS control, so the tuning results may be biased. Consequently, the \textbf{challenge (I)} of developing closed-loop DBS controllers is to ensure that the control policy can perform consistently over diverse 
and dynamic patient contexts and states.


Reinforcement learning (RL) has shown considerable potential in control over complicated systems~\cite{mnih2015human, gu2017deep, gaoreinforcement, gao2021gradient}, 
and various RL-based approaches have been proposed to facilitate closed-loop DBS~\cite{guez2008adaptive, pineau2009treating, nagaraj2017seizure, gao2020model}. Specifically, several approaches~\cite{guez2008adaptive, pineau2009treating, nagaraj2017seizure} model EEG and LFP as the state space of the RL environment and use temporal difference learning or fitted Q-iteration to 
design control policies adapting stimulation amplitudes/frequencies to conserve energy usage. The deep actor-critic based approach proposed in~\cite{gao2020model} further allows the temporal pattern of the stimuli to be adapted over time, benefiting from the use of deep RL techniques 
capable of searching in larger state and action space.
Although such methods achieve satisfactory control of efficacy and energy savings jointly, \emph{\textit{they have only been 
evaluated in simulations}}, \textit{i.e.}, on computational BG models~\cite{so2012relative,jovanov2018platform}. One may assume that unlimited training data can be obtained 
from such models, which is contrary to the real-world case where the device programming is done in clinics and the patient only participates sparsely over~time.

Another limitation of directly using deep RL methods for real-time 
DBS control is the computational complexity of evaluating the RL policies \textit{in vivo}, as they are usually represented by deep neural networks (DNNs) that may require millions of multiplications in a single forward pass. The resource-constrained implantable devices (\textit{e.g.}, Fig.~\ref{fig:Regions1mod}) may not support or facilitate such computations. Thus, the \textbf{challenge (II)} of closed-loop DBS is to ensure that the controller can be 
designed with limited training samples and executed without the need of extensive computing resources. 
Further, in contrast to simulated or robotic environments where most RL policies can be deployed directly 
for performance evaluation, the safety and control efficacy of the controllers 
directly used on patients need to be thoroughly 
evaluated before each test condition starts~\cite{parvinian2018regulatory}. Hence, the \textbf{challenge (III)} of 
enabling closed-loop DBS therapies in patients is being able to proactively provide accurate estimations of the expected performance of the controllers.

Consequently, in this paper, we first introduce an offline RL framework 
to address the challenges (I) and (II) above, resulting in a closed-loop DBS system that is both \textit{\emph{effective (in terms of therapy)}~and \emph{energy-efficient}}. 
Specifically, we model the BG regions of the brain as a Markov decision process (MDP), capturing the underlying neuronal activities in response to the stimuli. Then, the deep actor-critic algorithm~\cite{lillicrap2015continuous} is adapted to adjust the amplitude of the stimuli according to the changes in LFPs. A total of four patients, equipped with the Medtronic Summit RC+S DBS devices~\cite{stanslaski2018chronically}, participated in the data collection and testing trials in clinics. Given that the deep actor-critic framework is considered offline RL and can leverage all historically collected trajectories, \textit{i.e.}, experience replay to facilitate optimizing the control policy, we address challenge (I) 
by varying the level of activities, medications etc. of the patients before and during the trials. Similarly, experience collected from non-RL controllers can also be used to update the policy; for example, in the early stage of learning, a controller that generates uniformly random amplitudes (within some range) can 
facilitate exploring the state and action space. We also introduce model distillation/compression~\cite{hinton2015distilling} techniques specifically 
for the DBS systems, such that the RL policies can be captured by deep neural networks (DNNs) with significantly fewer nodes, whose forward passes can be executed 
within the required control rates, 
addressing challenge~(II). 

To address challenge (III), we introduce a model-based offline policy evaluation (OPE) method that captures the underlying dynamics of the considered MDP, where the expected returns of the control policy can be estimated by the mean return of the trajectories rolled out from the learned model, without directly deploying the policy to the patient. In each DBS trial, the control efficacy is evaluated from various sources, including LFP biomarkers recorded from the implantable DBS device, patient responses to bradykinesia tests, satisfaction level reported by the patient, and the overall tremor severity quantified from accelerometry data collected by external wearable devices (\textit{e.g.}, smart watch). Note that each of the latter three criteria is only evaluated once at the end of each~trial; yet they are imperative for evaluating the control efficacy from the patient's side. These efficacy metrics are thus considered sparsely available compared to the LFPs that can be 
sensed in each time step, which limits the use of  
existing OPE methods, including importance sampling (IS)~\cite{precup2000eligibility, gao2023hope}, distributional correction estimations (DICE)~\cite{nachum2019dualdice}, and the model-based OPE~\cite{gao2022offline}, as these do not allow for explicitly capturing/modeling such end-of-session rewards.
Our OPE method can capture such behaviors through a specially designed architecture and training objective, outperforming existing methods as we show in clinical experiments.

The contributions of this work are three-fold:
($i$) to the best of our knowledge, this is the first \textit{`full-stack'} offline RL 
methodology that facilitates both \textit{optimizing} and \textit{evaluating} RL-based DBS control policies using historical data; ($ii$) we developed an RL-based DBS controller whose performance 
is validated through clinical trials with PD patients, demonstrating \textit{reduced energy consumption with non-inferior control efficacy compared to cDBS} -- 
{\textbf{\textit{this is the first effective closed-loop DBS control that is not an ON/OFF switching, or scaling up/down proportionally, and has been extensively tested in clinic (i.e., on patients)}}}; ($iii$) our OPE method 
effectively captures the end-of-session rewards, leading to accurate estimations of control efficacy using the data collected in clinic; thus, 
helps demonstrate the 
effectiveness of the policies to be tested proactively, and can be used to prioritize the policies that could 
lead to better performance within the limited amount of testing~time.

This paper is organized as follows. 
Sec.~\ref{sec:prelim} provides the basics of DBS, RL, and OPE, before our clinical closed-loop DBS setup is introduced in Sec.~\ref{subsec:setup}.
In Sec.~\ref{sec:RL}, the offline RL framework is introduced, enabling training and updating RL controllers with historical data. Sec.~\ref{sec:OPE} introduces the model-based OPE approach to estimate performance of RL policies. Sec.~\ref{sec:eval} presents the results~of the experimental evaluations on patients, before concluding remarks in~Sec.~\ref{sec:discussion}.

%% file: Preliminary.tex
\section{Preliminaries and Motivation}
\label{sec:prelim}

In this section, we first introduce DBS, before presenting in the next section the DBS experimental setup 
we developed for clinical trials, including sensing, communication and control. Also, preliminaries for offline RL and OPE are briefly introduced; more comprehensive reviews of RL and OPE can be found in~\cite{gao2020model, lillicrap2015continuous, silver2014deterministic, gao2022offline}. 

\subsection{The Need for Closed-Loop DBS} 
\label{subsec:pd_dbs}
PD is caused by progressive death of dopaminergic neurons in the substantia nigra region of the brain. This change in dopaminergic signaling results in pathological activity in the BG regions targeted by DBS, \textit{globus pallidus pars interna} (\verb!GPi!), \textit{globus pallidus pars externa} (\verb!GPe!) and subthalamic nucleus (\verb!STN!); see Fig.~\ref{fig:Regions1mod}. Given the reduced number of neurons, the level of dopamine generally decreases in BG, leading to various motor symptoms such as bradykinesia and tremor~\cite{de2006epidemiology, kuhn2006reduction, brown2001dopamine}. Physiologically, the effect of PD can be captured by the changes in 
LFPs in \verb!GPi!, \verb!GPe! and \verb!STN!.
Specifically, PD can cause abnormal neuron firings in these regions, and lead to increased beta-band (13-35 Hz)
amplitude ($P_\beta$), referred to as the beta amplitude, of the LFPs~\cite{gao2022offline}. 

\begin{figure}[!t]
    \centering
    \includegraphics[width=.94\columnwidth]{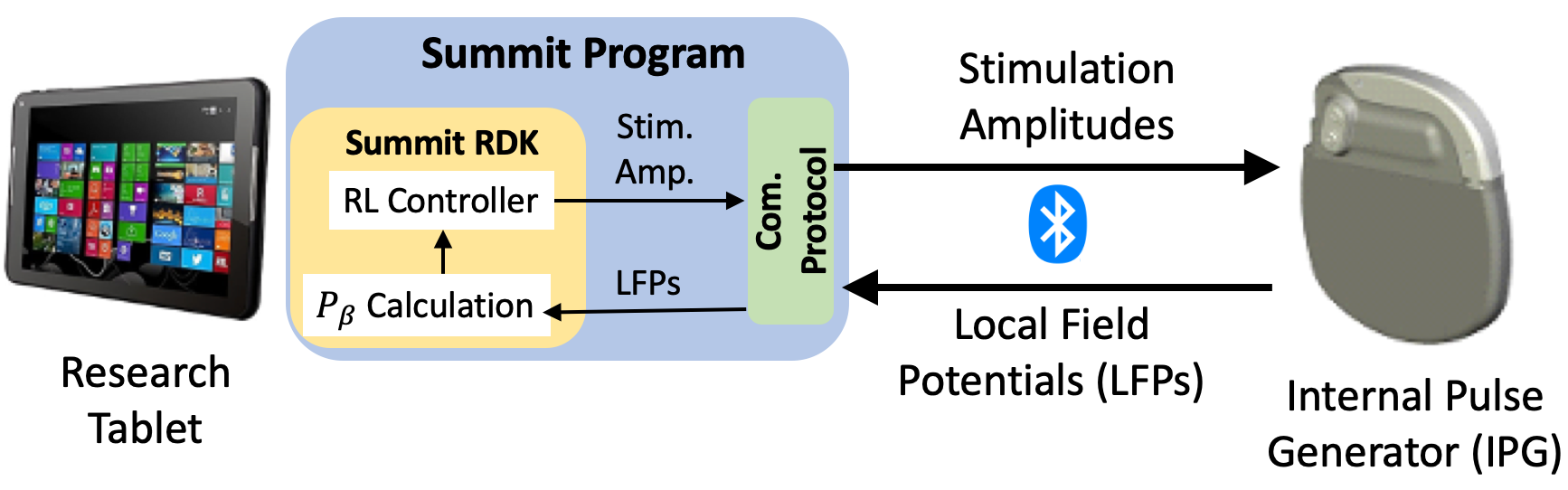}
    \vspace{-10pt}
    \caption{The overall architecture of the RC+S DBS system. The Summit research and development kit (RDK) can be used to configure the Summit program, allowing us to compute the beta amplitude ($P_\beta$) and execute the RL controller.}
    \vspace{-4pt}
     \label{fig:summit}
\end{figure}

Existing research-only DBS devices are capable of capturing the changes in LFPs through the multi-contact electrodes implanted in the BG. As illustrated in Fig.~\ref{fig:Regions1mod}, we used 4-contact electrodes placed in the \verb!STN! and \verb!GP! regions. Monopolar stimulation was delivered on a single contact on each lead (with the case serving as counter-electrode). The two contacts surrounding the stimulation contact were used for sensing LFPs (i.e., sandwich sensing).
%
Existing devices providing open-loop cDBS stimulate pulses at a fixed amplitude, which in most cases can correct the abnormal neuronal activity~\cite{kuncel2004selection}. However, constantly stimulating with high amplitudes significantly reduces the battery lifetime of the DBS device and may 
cause serious side-effects such as speech impairment~\cite{beric2001complications, little2013adaptive, swann2016gamma}. Consequently, it is important to design DBS controllers that are \textit{effective} (from the control, i.e., therapy,  perspective) and \textit{energy-efficient}.

As discussed in Introduction, current aDBS approaches 
require considerable time and effort for the patients and their healthcare providers to determine the thresholds through trial-and-error~\cite{wong2022proceedings}. 
Several deep-RL-based controllers have been proposed for closed-loop DBS, which can adapt the amplitude of the stimulation pulses in real time~\cite{gao2020model, gao2022offline} in response to changes in the feedback signals (\textit{e.g.}, $P_\beta$). However, such frameworks are only validated through numerical simulations, \textit{i.e.}, on \textit{simplified} computational BG models, instead of clinical trials with human participants. In real world, \textit{substantial} historical experience, or trajectories collected from past interactions between the controller and the environment (patient), may be necessary to learn an RL policy 
with suitable control efficacy and patient satisfaction~\cite{lee2020stochastic}. Offline RL 
holds promise to resolve this challenge, as it can use the data collected from any type of controllers, including cDBS or simply a policy switching between arbitrary stimulation amplitudes/frequencies, to optimize an RL control policy. Moreover, each time before a new control policy is deployed to the patient, the clinicians need to assess its 
effectiveness and may require justifications toward its estimated control efficacy and performance~\cite{parvinian2018regulatory}. OPE can facilitate such use cases, as it is capable of estimating the expected return of RL policies using historical trajectories, bridging the gap between the offline RL training and evaluations. Preliminaries for offline RL and OPE are presented in two subsections below.

\subsection{Offline Reinforcement Learning}
\label{subsec:offline_rl}

Offline RL has proven useful in many domains, including robotics~\cite{gu2017deep, qitong2019iccps}, healthcare~\cite{gao2021gradient}, etc., since it can optimize the control policies without requiring the environment to be presented, which guarantees the safety of the learning process. Further, it does not require the training data to be exclusively collected by the control policy being updated, leading to improved sample efficiency. To facilitate offline RL, the underlying dynamical environments are firstly modeled as Markov decision processes (MDPs).

\begin{definition}[MDP]\label{def:mdp}
An MDP is a tuple $\mathcal{M} = (\mathcal{S}, s_0, \mathcal{A}, \mathcal{P}, R, \gamma)$, where $\mathcal{S}$ is a finite set of states; $s_0$ is the initial state; $\mathcal{A}$ is a finite set of actions; $\mathcal{P}$ is the transition function defined as $\mathcal{P} : \mathcal{S} \times \mathcal{A} \rightarrow \mathcal{S}$; $R: \mathcal{S} \times \mathcal{A} \times \mathcal{S} \rightarrow \mathbb{R}$ is the reward function, and $\gamma \in [0, 1)$ is a discount factor.
\end{definition}

Then, the RL policy $\pi: \mathcal{S} \rightarrow \mathcal{A}$ determines the action $a=\pi(s)$ to be taken at a given state $s$. The accumulated return under a policy $\pi$ can be defined as follows.

\begin{definition}[Accumulated Return]\label{def:rl_return}
Given an MDP $\mathcal{M}$ and a policy $\pi$, the accumulated return over a finite horizon starting from the stage $t$ and ending at stage $T$, for $T>t$, is defined as 
\begin{align}
\label{eq:acc_return}
G_t^\pi = \sum\nolimits_{k=0}^{T-t} \gamma^{t+k} r_{t+k},    
\end{align}
where $r_{t+k}$ is the return at the stage $t+k$.
\end{definition}

The goal of offline RL can now be defined as follows.

\begin{problem}[Offline Reinforcement Learning]
\label{prob:rl_objective}
Given an MDP $\mathcal{M}$ with \underline{unknown transition dynamics} $\mathcal{P}$, a pre-defined reward function $R$, and a experience replay buffer $\mathcal{E}^\mu =\{[(s_0,a_0,r_0,s_1),\dots,$ $(s_{T-1},a_{T-1},r_{T-1},s_T)]^{(0)}, [(s_0,a_0,r_0,s_1),\dots]^{(1)}, \dots|a_{t}\sim\mu(a_{t}|s_{t})\}$ containing trajectories collected over an \underline{unknown behavioral policy} $\mu$, find the target policy $\pi^*$ such that the expected accumulative return starting from the initial stage over the entire horizon is maximized,~\textit{i.e.}, 
\begin{align}
\label{eq:pi_objective}
\pi^*  = \argmax_\pi \mathbb{E}_{s,a\sim\rho^\pi,r\sim R}[G_0^\pi]; 
\end{align}
here, $\rho^\pi$ is the state-action visitation distribution under policy $\pi$.
\end{problem}

The deep actor-critic RL framework~\cite{lillicrap2015continuous} can be leveraged to solve~\eqref{eq:pi_objective}. 
Other value-based RL methods such as conservative Q-learning~\cite{kumar2020conservative} and implicit Q-learning~\cite{kostrikov2021offline} could  also be considered; however, actor-critic methods can in general reduce the variance of gradient estimations and result in faster convergence~\cite{mnih2016asynchronous, wu2017scalable,gao2020model}. Here, we specifically consider the deterministic version of actor-critic~\cite{lillicrap2015continuous}, instead the one producing stochastic policies~\cite{haarnoja2018soft}, as it would be easier to demonstrate the 
effectiveness of deterministic policies in clinics, as well as 
via OPE methods introduced~below. {Details on the deep actor-critic algorithm~\cite{lillicrap2015continuous} are provided in Appendix~\ref{app:deep_ac}.}




\subsection{Offline Policy Evaluation for DBS}
\label{subsec:ope_prelim}

\begin{figure}[!t]
    \centering
    \includegraphics[width=1.0\columnwidth]{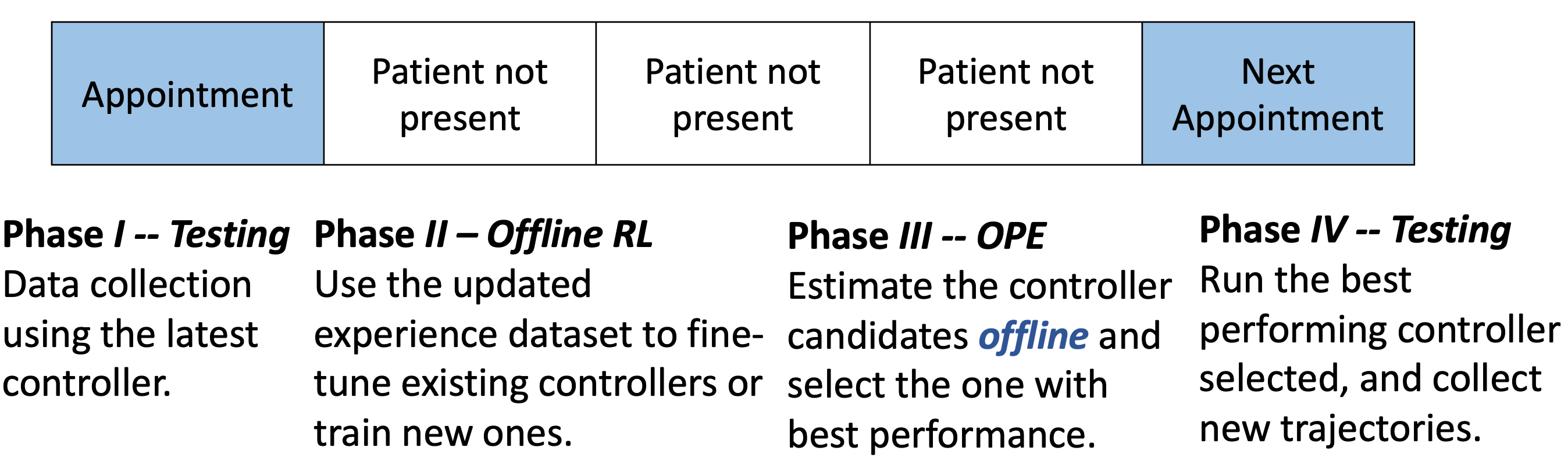}
    \vspace{-14pt}
    \caption{Timeline for training RL-based~DBS controllers in clinical studies. Since only limited~data can be collected during each clinical visit, offline RL can be used to fine-tune existing or train new controllers using all the historical data. Then, offline policy evaluation (OPE) facilitates choosing the possible top-performing ones to be tested in the next visit.
    }
    \label{fig:ope_rl}
\end{figure}

OPE allows the use of experience replay buffer to estimate the expected return of RL policies, without the need of deploying them to the environment directly. Fig.~\ref{fig:ope_rl} illustrates the use case of OPE in the context of DBS clinical testing. Specifically, during phase $I$ and $II$, offline RL uses all trajectories collected historically to train RL policies following different hyper-parameters etc. Then, in phase $III$, OPE can be used to estimate and rank the expected return of these policies, where the top-performing ones can be deployed during the next clinic visit (phase $IV$). Consequently, 
OPE can effectively reduce the number of testing sessions needed, so the policies that show promise attaining better performance can be thoroughly tested within the short time frame. Also, it can 
demonstrate 
the effectiveness of the policies to be deployed~in~clinics. 

The goal of OPE can be defined as follows.
\begin{problem}[Offline Policy Evaluation]
\label{prob:ope_objective}
Consider a \underline{target} policy $\pi$, and \underline{off-policy} trajectories $\mathcal{E}^\mu = \{(s_0, a_0),(s_1$ $,a_1),\dots|a_t=\mu(s_t)\}$, collected following a \underline{behavioral} policy $\mu \neq \pi$, over an MDP $\mathcal{M}$.
The OPE goal 
is to estimate the expected return of the target policy~$\pi$, \textit{i.e.}, $\mathbb{E}_{s,a\sim\rho^\pi,r\sim R}[G_0^\pi]$. 
\end{problem}

Most existing OPE methods, such as~\cite{precup2000eligibility, gao2023hope, liu2018breaking, dai2020coindice, jiang2016doubly, thomas2016data, yang2020off, tang2019doubly, fu2020benchmarks}, are heavily based on importance sampling (IS) and could result in inconsistent estimations due to the high variance of the IS weights~\cite{liu2018breaking, dai2020coindice}. On the other hand, model-based OPE methods have shown strengths in estimating the expected returns more accurately~\cite{fu2020benchmarks, gao2022offline}, by directly capturing the MDP transitions and rewards. The variational encoding-decoding based deep latent MDP model (DLMM) introduced in~\cite{gao2022offline} is shown to be effective evaluating 
control
policies for a computational BG model. Specifically, DLMM is 
derived following the variational inference framework from~\cite{kingma2013auto}. 
{The basics of DLMM are provided in Appendix~\ref{app:dmll}, }
and we refer the readers to~\cite{kingma2013auto} for basics of variational inference. In Sec.~\ref{sec:OPE}, we extend it toward 
the clinical use case considered in this work, to allow for including the QoC metrics that can be only evaluated once in each session, such as the bradykinesia results, patient ratings, and tremor severity, which will be available as illustrated~in~Fig.~\ref{fig:clinical_setup}.

\section{DBS Setup Used in Clinical Trials}
\label{subsec:setup}

\begin{figure}[!t]
    \centering
    \includegraphics[width=.88\columnwidth]{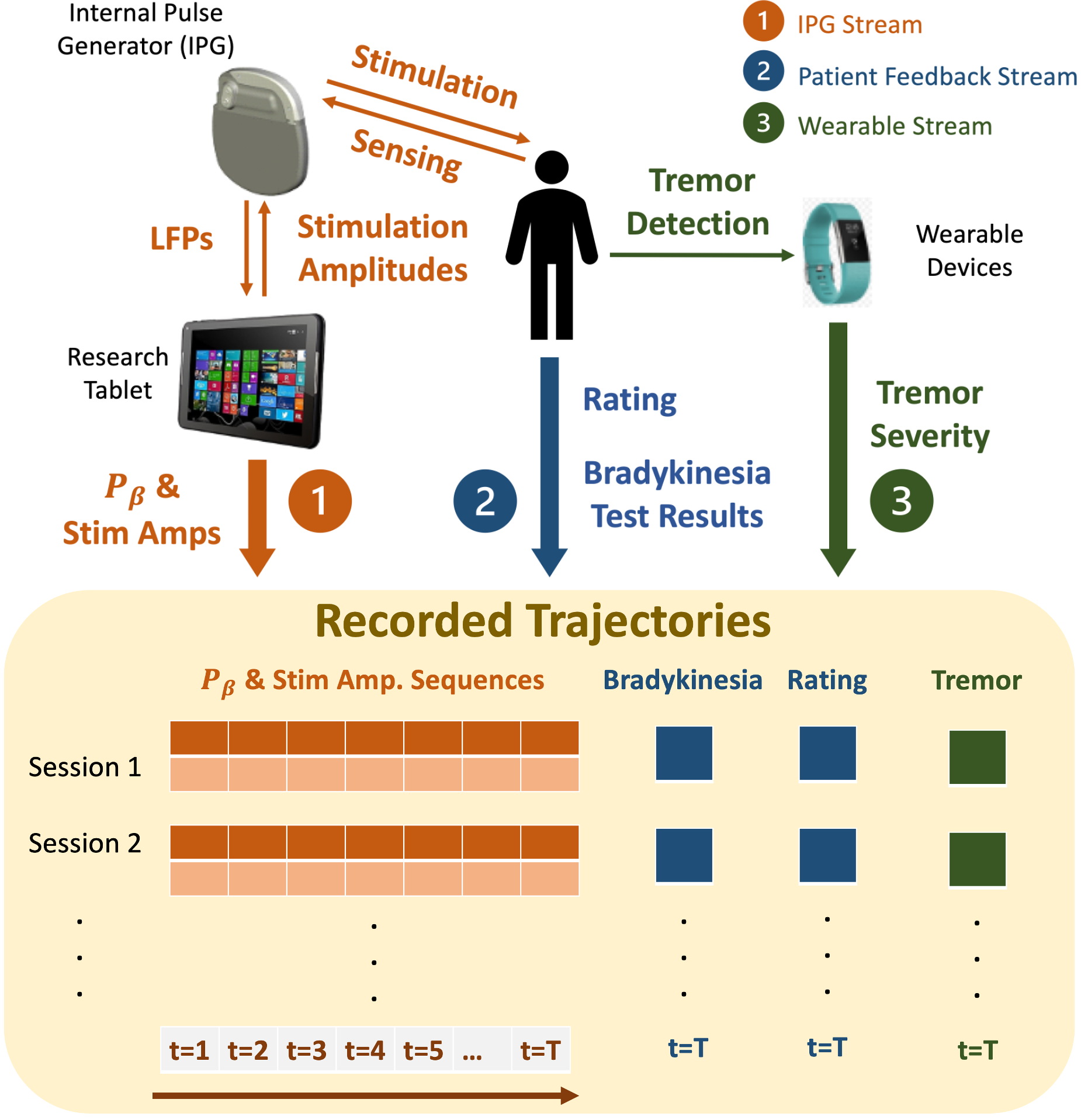}
    \vspace{-8pt}
    \caption{Setup of the developed DBS clinical testing procedure. A total of three data streams 
    are collected: 
    ($1$)~the LFPs and stimulation amplitudes are recorded over time; the logged trajectories 
    are used to evaluate the performance of deployed RL controllers, 
    as well as training data for further fine-tuning; ($2$)~patient feedback including results from bradykinesia tests and a rating on the scale between 1-10; ($3$)~patient tremor severity captured by wearable devices.
    }
    \label{fig:clinical_setup}
     \vspace{-2pt}
\end{figure}

We build on the research-only Medtronic's Summit RC+S system~\cite{stanslaski2018chronically} 
to enable testing of RL-based controllers in clinical trials. The overall architecture of the RC+S-based system we developed is illustrated in Fig.~\ref{fig:summit}. Specifically, 
Medtronic provides the code 
and communication 
APIs (Summit program), 
which enable the stimulation amplitude of the pulses delivered by the internal pulse~generator (IPG) to be adapted over time. The Summit program is developed using the \texttt{C\#} language under the \texttt{.NET} framework,  
which we extended to execute RL policies leveraging the provided Summit research development kit (RDK), requiring the use~of~a~Windows~OS.

Thus, a research tablet
is used for the execution of the 
developed DBS controllers; 
the desired stimulation amplitude is computed for each control cycle (every 2 seconds) and sent to the IPG over Bluetooth$^\text{TM}$, using 
proprietary communication and security protocols. On the other hand, the IPG 
transmits to the controller the LFPs captured from the BG, from which the beta amplitude of the LFPs, denoted by $P_\beta$, is calculated and used as a quality of control (QoC) metric as well as potential control feedback signals (\textit{i.e.}, inputs to the RL controller). Each clinical trial session lasts 5-20 minutes depending on the schedule of the visit, and multiple controllers can be tested across different sessions. All the computed $P_\beta$ and stimulation amplitudes applied over time are logged for future training and evaluation purposes, as summarized 
in Fig.~\ref{fig:clinical_setup}. 
For the developed system design, we obtained the FDA's Investigative Device Exception (IDE) G180280, which has allowed us to perform human experiments according to an Institutional Review Board (IRB) protocol approved by Duke University Medical Center.

In addition to $P_\beta$, three other QoC metrics are collected from every patient at the end of each session. Specifically, near the end of each session, the patient is asked to perform 10 seconds of hand grasps (rapid and full extension and close of all fingers) maneuver~\cite{ramaker2002systematic} to evaluate the severity of the possible bradykinesia caused by PD. Such hand motions are captured by a leap motion sensor 
by Ultraleap~\cite{butt2018objective}. Then, the elapsed time between any two consecutive open fist is captured and recorded by the sensor, after which the grasp frequency can be calculated as 
\begin{align}
\label{eq:grasp_speed}
    QoC_{grasp} = \frac{1}{\frac{1}{N-1}\sum_{i=1}^{N-1} t_{(i, i+1)}};
\end{align}
here, $N$ is the total number of open fists 
throughout the 10~s test, and $t_{(i,i+1)}$ is the time spent between the $i$-th and $i+1$-th grasp.~Further, at the end of each session, the patient provides a score between 1-10, with 10 indicating the highest level of satisfaction with the treatment received in the past session, and 1 being the lowest,~\textit{i.e.},
\begin{align}
\label{eq:patient_rating}
    QoC_{rate} \in [1,10] \subset \mathbb{Z}^+ .
\end{align}
The grasp frequency and rating for each session are also recorded, which corresponds to the patient feedback stream in Fig.~\ref{fig:clinical_setup}.

Throughout all sessions, an Apple watch is worn by the patient at their wrist, where the Apple's movement disorders kit~\cite{powers2021smartwatch} is used to analyze the accelerometry movements, classifying the patient's tremor severity as no-tremor, slight, mild, moderate and strong every 1 minute, following StrivePD's implementation~\cite{chen2021role}. At the end of each session, an overall tremor severity is recorded as the fraction of time the patient experiencing mild ($T_{mild}$), moderate ($T_{moderate}$) or strong ($T_{strong}$) tremor over the entire session with length $T_{session}$,~\textit{i.e.},
\begin{align}
\label{eq:tremor_severity}
    QoC_{tremor} = \frac{T_{mild} + T_{moderate} + T_{strong}}{T_{session}} \times 100\%.
\end{align}

The three data streams are collected from all trial sessions after each clinical visit. 
Moreover, each time a patient may come into the clinic with slightly different PD conditions (\textit{e.g.,} pathology progression over time), medication prescriptions, activity levels etc.; thus, 
our goal is to capture impact of such changes by the data collection process, 
in order to facilitate the training and testing the offline RL and 
OPE frameworks for DBS.

%% file: RL.tex
\section{Offline RL Design of DBS Controllers} 
\label{sec:RL}

In this section, we 
employ offline RL for learning control policies for DBS clinical trials, 
starting from the formulation of an MDP $\mathcal{M}$ capturing the underlying neurological dynamics in the BG, and the policy distillation technique that allows for reducing the computational time and resource needed to evaluate the RL policies (represented by DNNs).

\subsection{Modeling the BG as an MDP}
\label{subsec:mdp}

We now define the elements of an MDP $\mathcal{M} = (\mathcal{S}, s_0, \mathcal{A}, \mathcal{P}, R, \gamma)$. 

\paragraph{State Space $\mathcal{S}$ and the Initial State $s_0$}
As discussed in Sec.~\ref{subsec:pd_dbs} and~\ref{subsec:setup}, our DBS controller supports calculation of $P_\beta$ from LFPs, and the changes in $P_\beta$ can be used 
as a biomarker for PD-levels for some patients.
Thus, we consider the MDP state, at a \textit{discrete} time step $t$, as a historical sequence of $P_\beta$ sampled at a fixed intervals, captured by $m \in \mathbb{Z}^+$, over a sliding queue of size $W \in \mathbb{Z}^+$,~\textit{i.e.},
\begin{align}
 s_t = \begin{bmatrix}
     \beta_{(\tilde t - (W-1)m)}, \beta_{(\tilde t- (W-2)m)}, ..., \beta_{(\tilde t-2m)}, \beta_{(\tilde t-m)}, \beta_{(\tilde t)}
    \end{bmatrix}.
\label{eq:mdp_s}
\end{align}
Here, $\beta_{(\tilde t)}$'s are the $P_\beta$ evaluated at the elapsed time $\tilde t$ since the clinical trial starts, $m$ is configurable in our system design 
(Fig.~\ref{fig:summit}), and we used $m=2$ corresponding to calculating $P_\beta$ every $2~s$, resulting in  $20~s$ time-windows for $W=10$ elements in the queue; finally,  $s_t \in \mathbb{R}^W$ is the state at $t$-th (discrete) step 
of the MDP.
The initial state $s_0$ is considered to be the $\beta$ sequence collected right before the clinical trial starts, \textit{i.e.}, from $\tilde t=-(W-1)m$ to $\tilde t = 0$.

\paragraph{Action Space $\mathcal{A}$}
The amplitude of DBS stimulation pulses can be changed in pre-defined (discrete) time steps, \textit{i.e.}, every 2 seconds 
for the developed controllers. We consider the actions $a_t$ as the 
percentage of the cDBS amplitude determined by clinicians; \textit{i.e.}, $a_t \in [0,1] \subset \mathbb{R}$, where $a_t=0$ and $a_t=1$ correspond to no-DBS and stimulation with the same amplitude as in cDBS, respectively.

\paragraph{Transition Dynamics $\mathcal{P}: \mathcal{S}\times\mathcal{A}\rightarrow\mathcal{S}$} 
Every time after the stimulation amplitude is adjusted following $a_t$, the system computes the latest $\beta_{(\tilde t + m)}$ using the LFPs sent back from the IPG; this leads to the MDP state at the (t+1)-th (discrete) step as
\begin{align}
    s_{t+1} = \begin{bmatrix}
     \beta_{(\tilde t-(W-2)m)}, \beta_{(\tilde t-(W-3)m)}, \dots, \beta_{(\tilde t)}, \beta_{(\tilde t+m)}
    \end{bmatrix}, \label{eq:mdp_s_2}
\end{align}
\textit{i.e.}, the left-most element in~\eqref{eq:mdp_s} is pushed out, with $\beta_{(\tilde t + m)}$ appended to the right-end. Note that we define the MDP states $s_t$ and actions $a_t$ over discrete time steps, $t$'s, instead the elapsed time $\tilde t$, for the conciseness of equations and presentations below. 
Now, the MDP transitions are captured to directly follow $s_{t+1} \sim \mathcal{P}(s_t, a_t)$. 

\paragraph{Reward Function $R:\mathcal{S}\times\mathcal{A}\rightarrow\mathbb{R}$}
Following from the setup of the DBS system (Sec.~\ref{subsec:setup}), we define the rewards as 
\begin{align}
\label{eq:per_step_reward}
    R(s_t,\hspace{-1pt}a_t,\hspace{-1pt}s_{t+1}) \hspace{-2pt}= \hspace{-2pt}\left \{
    \begin{array}{ll}
        r_a - C_1 \cdot a_t, &\text{if }  \bar \beta_{(\tilde t + m)} < \xi_\beta;\\
        r_b - C_1 \cdot a_t, &\text{if } \bar \beta_{(\tilde t + m)} \geq \xi_\beta;
    \end{array}
    \right . 
\end{align}
specifically, if the beta amplitude received at the $(t+1)$-th step, $\beta_{(\tilde t + m)}$, is less than some threshold $\xi_\beta$, then a non-negative reward $r_a$ is issued along with the term $-C_1 \cdot a_t$ ($C_1>0 $, $C_1\in \mathbb{R}$) penalizing over-usage of large stimulation amplitudes (for better energy efficiency). On the other hand, if $\beta_{(\tilde t + m)}$ is greater than the threshold $\xi_\beta$, a negative reward $r_b$ will be used to replace $r_a$ above. 
\begin{remark}
\label{remark1}
The reward functions used for RL training do not consider the QoC metrics that are available not at every step of the control execution (i.e., every $2~s$) but only at the end of each clinical session, \textit{i.e.}, $QoC_{grasp}, QoC_{rate}, QoC_{tremor}$ from~\eqref{eq:grasp_speed}, \eqref{eq:patient_rating}, \eqref{eq:tremor_severity}. The reason is that the horizon $T$ is usually large and the their coverage can be very sparse. Instead, these QoC metrics serve as great measurements quantifying how well the policies perform, which are thus leveraged by the OPE techniques introduced in Sec.~\ref{sec:OPE}.
\end{remark}

For the introduced MDP $\mathcal{M}$, we
leverage the offline RL framework introduced in Sec.~\ref{subsec:offline_rl} to search for the target policy $\pi^*$. Following from Problem~\ref{prob:rl_objective}, it requires an experience replay buffer $\mathcal{E}^\mu$ that consists of historical trajectories collected over some behavioral policy $\mu$. At the beginning of offline RL training, exploration of the environment is deemed more important than exploitation~\cite{ishii2002control}. Hence, a controller that generates random actions uniformly from $[B, 1]$ is used to constitute $\mathcal{E}^\mu$ at earlier stage of clinical trials, where $B$ is the lower bound from which the random $a_t$ can be generated, for the sake of patient's safety and acceptance.

Once the RL policies can attain satisfactory overall performance,~\textit{i.e.}, quantified as achieving significantly improved QoCs (introduced in Sec.~\ref{subsec:setup}) compared to the random controller above, we consider including into $\mathcal{E}^\mu$ the trajectories obtained from such RL policies. From this point onward, the replay buffer $\mathcal{E}^\mu$ will be iteratively updated and enriched with the RL-induced trajectories after each trial. Consequently, the behavioral policy $\mu$ can be considered as a mixture of random control policy and several RL policies deployed in past trials in general. With $\mathcal{E}^\mu$ being defined, the objective for training RL policies,~\eqref{eq:ac_objective}, can be optimized using gradient descent~\cite{lillicrap2015continuous, gao2020model, gao2022offline}.

\subsection{Policy Distillation}
\label{subsec:distillation}

Our system design (Fig.~\ref{fig:summit}) is set to process various tasks in each $2~s$ stimulation (i.e., control) period, 
facilitating communication between the research tablet and IPG, computing $P_\beta$ from LFPs, evaluating the RL controller, data logging, and other basic functionalities that ensure the safety and functionality of DBS. Hence, it was 
critical to reduce the overall computation requirements, 
such that each task meets the required timings, 
as well as prolong the battery lifetime. 
As introduced in Sec.~\ref{subsec:offline_rl}, the RL policies are parameterized as DNNs; although a forward pass of a DNN would not require as much computational resources as for training (through back-propagation), it may still involve hundreds of thousands of multiplication operations. For example, consider the recommended DNN size as  in~\cite{lillicrap2015continuous}, it takes at least 120,000 multiplications to evaluate a two-layer NN with 400 and 300 nodes each. 
Hence, 
we integrate into our system the model/policy distillation techniques~\cite{hinton2015distilling}, allowing smaller sized NNs to be used to parameterize RL policies.

We build on a similar approach as in~\cite{rusu2016policy}, 
originally proposed to reduce the size of DNNs used in deep Q-learning~\cite{mnih2015human}, which only works for a discrete action space. In particular, our extension allows for the use in the deterministic actor-critic cases considered in this work. Consider the original policy (\textit{teacher}) $\pi_{\theta_a}$ parameterized by a DNN with weights $\theta_a$. We train a smaller-sized DNN (\textit{student}) with weights $\tilde \theta_a$ to learn $\theta_a$'s behavior, by minimizing the mean squared~error 
\vspace{-6pt}
\begin{align}
\label{eq:distillation}
    \min_{\tilde \theta_a} || \pi_{\theta_a}(s_t) - \pi_{\tilde \theta_a}(s_t) ||^2 ,
\end{align}
for all state samples contained in the experience replay $ s_t \in \mathcal{E}^\mu$. We also consider augmenting the data used to optimize~\eqref{eq:distillation} to smooth out the learning process. We introduce synthetic states, $\tilde s_t$'s, where each $\tilde s_t$ is generated by adding noise 
to each dimension of a state sample $s_t$ that is originally in $\mathcal{E}^\mu$;
the noise is 
sampled from a zero-mean Gaussian distribution, $\epsilon_t \sim \mathcal{N}(0, \sigma^2)$ with $\sigma$ being a hyper-parameter.

%% file: OPE.tex


\section
{OPE of DBS Controllers Including Patient Feedback and Tremor Data}
\label{sec:OPE}

As discussed in Remark~\ref{remark1}, besides the reward function introduced in Sec.~\ref{subsec:mdp}, for OPE we employ QoC metrics $QoC_{grasp}$, $QoC_{rate}$, and $QoC_{tremor}$ defined in \eqref{eq:grasp_speed},~\eqref{eq:patient_rating},~\eqref{eq:tremor_severity}, respectively, which are only available at the end of each session. As these 
well-capture performance (i.e., therapy effectiveness) of the considered policy, 
for OPE we additionally consider the end-of-session rewards 
defined~as 
\begin{align}
\label{eq:end_of_session_reward}
    r_{end} = & ~R_{end}(s_0, a_0, s_1, a_1, ..., s_{T-1}, a_{T-1}, s_T) \nonumber\\= & ~C_2 \cdot QoC_{grasp} + C_3 \cdot QoC_{rate} - C_4 \cdot QoC_{tremor},
\end{align}
with 
$C_2$, $C_3$, $C_4>0$ real constants. 
Without loss of generality, we slightly modify the 
total return under policy~$\pi$ (from Problem~\ref{prob:ope_objective}) as 
\begin{align}
    \label{eq:ope_total_return}
    G_0^\pi = r_{end} + \sum\nolimits_{t=0}^T \gamma^t r_t,
\end{align}
where $r_t$ and $r_{end}$ follow from~\eqref{eq:per_step_reward} and~\eqref{eq:end_of_session_reward}, respectively.

\begin{figure}[t!]
    \centering
    \includegraphics[width=.82\linewidth]{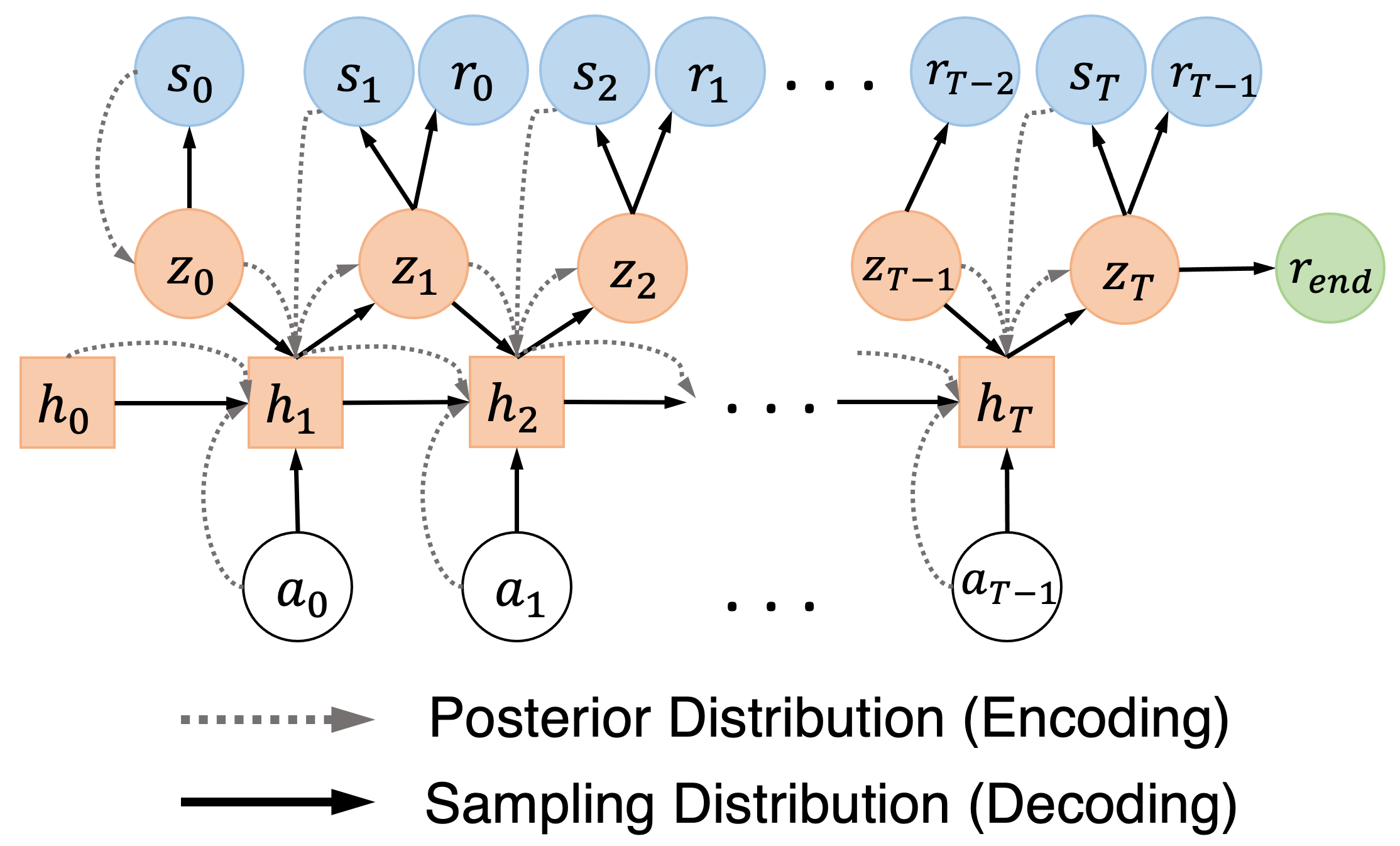}
    \vspace{-6pt}
    \caption{Architecture of the new deep latent sequential model (DLSM). The conditional dependencies between the variables from the posterior and sampling distributions are shown in dashed and solid lines, respectively.}
    \label{fig:vlsm}
\end{figure}

As discussed in Sec.~\ref{subsec:ope_prelim}, the DLMM introduced in~\cite{gao2022offline}, falls short in dealing with long horizons and predicting the end-of-session rewards $r_{end}$. To address these limitations, in this section 
we introduce the \textit{deep latent sequential model} (DLSM) that directly enforces the transitions over the LVS. The overall model architecture is shown in Fig.~\ref{fig:vlsm}. First, the latent prior $p_\psi(z_0)$ is defined only over the initial latent variable at step $t=0$, $z_0$, which follows a multivariate Gaussian distribution with zero mean and identity covariance~matrix. 

Then, the encoder (approximated posterior) is defined over each trajectory (from $t=0$ to $T$) as
\vspace{-8pt}
\begin{align}
\small
    & q_\phi(z_{0:T}|s_{0:T},a_{0:T-1})  =  q_\phi(z_0|s_0) \prod\limits_{t=1}^{T} q_\phi(z_t|z_{t-1},a_{t-1},s_t).
\end{align}
Further, the second term $q_\phi(z_t|z_{t-1},a_{t-1},s_t)$, which enforces the transitions between $z_{t-1}$ and $z_t$ conditioned on $(a_{t-1}$, $s_t)$ and enables the encoder to capture the dynamical transitions in the LVS, can be obtained iteratively following
\begin{align}
\label{eq:encoder}
\small
    z_0^\phi  \sim q_\phi(z_0|s_0), \;
    h_t^\phi  = f_\phi(h_{t-1}^\phi, z_{t-1}^\phi, a_{t-1}, s_{t}),  \;
    z_t^\phi  \sim q_\phi(z_t|h_t^\phi);
\end{align}
here, $q_\phi(z_0|s_0)$ and $q_\phi(z_t|h_t^\phi)$ are parameterized by multivariate diagonal Gaussian distributions, each with mean and covariance determined by a feedforward DNN~\cite{bishop2006pattern}; moreover, $h_t^\phi$ is the hidden state of a recurrent DNN, such as long short-term memory (LSTM)~\cite{hochreiter1997long}, capturing the historical transitions among $s_t$, $a_t$ and $z_t^\phi$ for all past steps up until $t-1$ within each trajectory.

The decoder (sampling distribution) is responsible for interacting with the target policies to be evaluated, from which the expected returns can be estimated as the mean return obtained by the simulated trajectories. Specifically, the decoder is defined as follows,~\textit{i.e.},
\begin{align}
\small
    p_\psi(z_{1:T}, s_{0:T}, & r_{0:T-1}, r_{end}|z_0) =  p_\psi(r_{end}|z_T) \cdot \nonumber\\ & \prod\limits_{t=0}^{T} p_\psi(s_t|z_t) \prod\limits_{t=1}^{T} p_\psi(z_t|z_{t-1}, a_{t-1})p_\psi(r_{t-1}|z_t);
\end{align}
here, $p_\psi(r_{end}|z_T)$ estimates the end-of-session rewards given the latent variable at $t=T$, $z_T$; $p_\psi(s_t|z_t)$, $p_\psi(r_{t-1}|z_t)$ reconstruct the states and rewards; $p_\psi(z_t|z_{t-1}, a_{t-1})$ enforces the transitions over the latent variables, $z_t$'s, conditioned on the actions; and $z_0 \sim p_\psi(z_0)$ is sampled from the prior. As a result, each simulated trajectory can be generated by the decoder following
\begin{align}
\label{eq:decoder}
\small
    & h_t^\psi  = f_\psi(h_{t-1}^\psi, z_{t-1}^\psi, a_{t-1}), \;
    z_t^\psi  \sim p_\psi( z_t| h_t^\psi), \; 
    s_t^\psi \sim p_\psi(s_t|z_t^\psi), \nonumber\\
    & r_{t-1}^\psi  \sim p_\psi( r_{t-1}|z_t^\psi),\;
    a_{t-1} \sim \pi(a_{t-1}|s_{t-1}^\psi),\;
    r_{end}^\psi \sim p_\psi(r_{end}|z_T);
\end{align}
here, $h_t^\psi$ is the hidden state of a recurrent DNN; $p_\psi(z_t| h_t^\psi)$, $p_\psi(s_t|z_t^\psi)$, $p_\psi( r_{t-1}|z_t^\psi)$ and $p_\psi(r_{end}|z_T)$ are multivariate diagonal Gaussians with means and covariances determined by four feedforward DNNs separately. Hence, $s_t^\psi$'s and $r_{t-1}^\psi$'s can be sampled iteratively following the process above, using the actions obtained from the target policy $a_{t-1} \sim \pi(a_{t-1}|s_{t-1}^\psi)$ accordingly, which constitute the simulated trajectories; and $r_{end}^\psi$ is sampled at the end of each simulated~trajectory.

The theorem below derives an ELBO for the joint log-likelihood $\log p_\psi(s_{0:T}, r_{0:T-1}, r_{end})$, following the above DLSM architecture.

\begin{theorem}[ELBO for DLSM]
\label{thm}
An ELBO of the joint log-likelihood $\log p_\psi(s_{0:T}, r_{0:T-1}, r_{end})$ can be obtained as
\begin{align}
\small
     \mathcal{L}&_{ELBO} (\psi,\phi) =  \mathbb{E}_{z_t\sim q_\phi} \Big[\sum\nolimits_{t=0}^T \log p_\psi(s_t|z_t) + \sum\nolimits_{t=1}^T \log p_\psi(r_{t-1}|z_t) \nonumber\\ & + \log p_\psi(r_{end}|z_T) - KL\big(q_\phi(z_0|s_0) || p(z_0)\big)  \nonumber\\
    & - \sum\nolimits_{t=1}^T  KL\big(q_\phi(z_t|z_{t-1},a_{t-1},s_t)||p_\psi(z_t|z_{t-1},a_{t-1})\big)  \Big] \label{eq:elbo} \\
    \leq & \log p_\psi(s_{0:T}, r_{0:T-1}, r_{end});
\end{align}
here, the first three terms are the log-likelihood of the decoder to reconstruct $s_t$, $r_{t-1}$ and $r_{end}$ correctly, and the two terms that follow regularize the transitions captured by the encoder over the LVS, with $KL(\cdot||\cdot)$ being the Kullback–Leibler (KL) divergence~\cite{kullback1951information}.
\end{theorem}
The proof of Theorem~\ref{thm} can be found in Appendix~\ref{app:proof}. 
Empirically, similar to the DLMM~\cite{gao2022offline}, {the ELBO can be evaluated using the trajectories from the experience replay $\mathcal{E}^\mu$, by replacing the expectation as the mean over all trajectories, after which the objective $\max_{\psi,\phi}\mathcal{L}(\psi,\phi)$ can be achieved} using gradient descent~\cite{kingma2014adam} following the algorithm in Appendix~\ref{app:alg}. Moreover, the reparameterization trick~\cite{kingma2013auto} is used, which allows for the gradients to be back-propagated when sampling from Gaussian distributions with means and covariances determined by DNNs. {Details on reparameterization can be found in~\cite{gao2022offline, kingma2013auto}.}


%% file: Evaluation.tex

\begin{figure*}[t!]
    \centering
    \includegraphics[width=.984\linewidth]{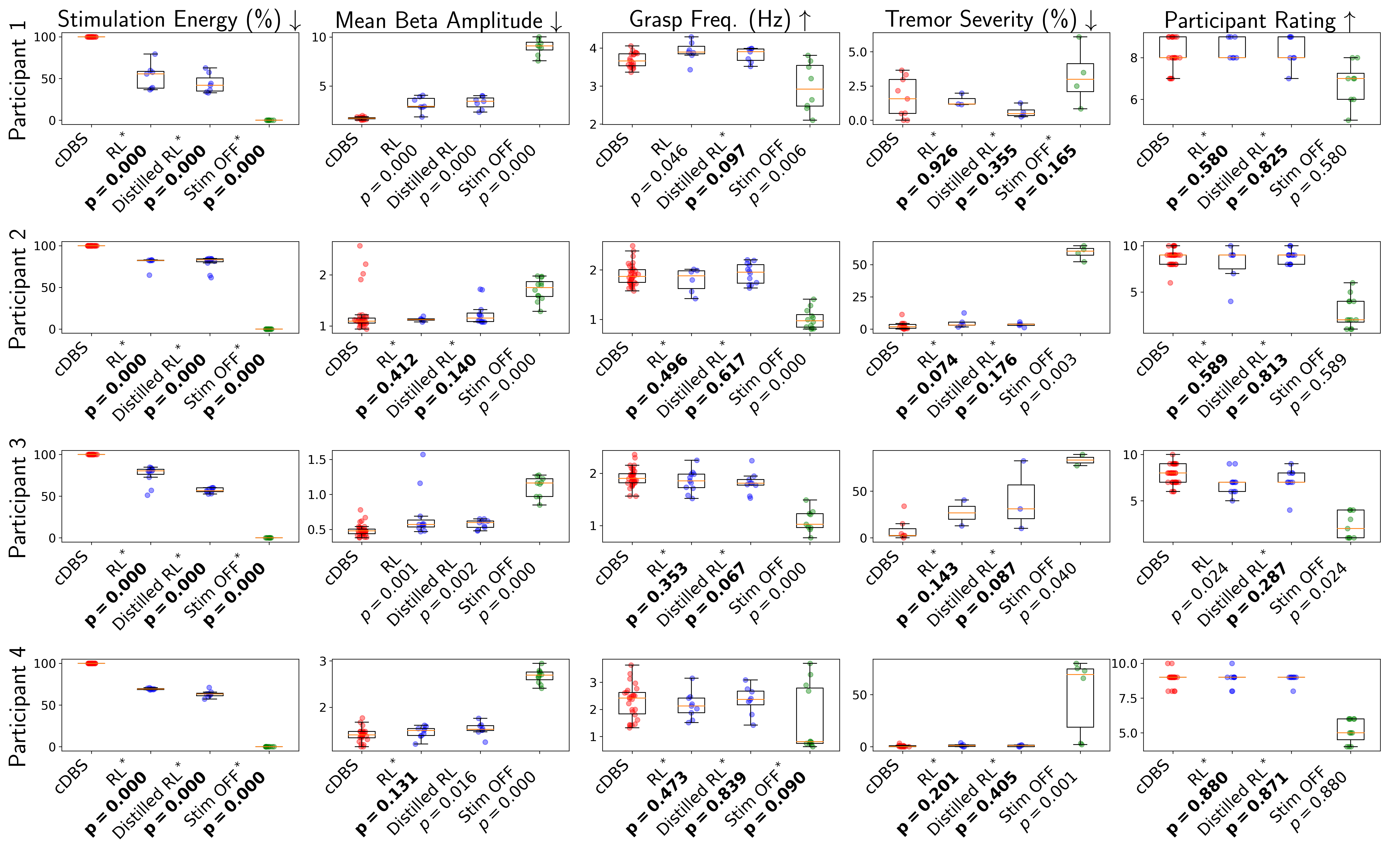}
    \vspace{-12pt}
    \caption{Quality of control (QoC) results from all clinical trials across participants. Wilcoxon rank-sum tests~\cite{mann1947test} between cDBS and each of the other controllers are used to test the null hypothesis that two sets of measurements are drawn from the same distribution, resulting in the $p$-values reported above. The null hypothesis is rejected when consider the stimulation energy consumed by both RL controllers, illustrating that they lead to significant energy reduction compared to cDBS. For all other QoCs, the null hypothesis is accepted in majority cases, showing that both RL controllers can in general attain similar control efficacy to cDBS. The controllers that lead to the acceptance/rejection of the null hypothesis in the desired direction are highlighted with asterisks and bold $p$-values.}
    \label{fig:overall_results}
    \vspace{-4pt}
\end{figure*}

\section{Clinical Evaluations}
\label{sec:eval}

Using our closed-loop DBS system presented in Sec.~\ref{subsec:setup}, we evaluated the developed RL-based control framework in clinical trials on four PD patients, {at Duke University Medical Center}. In particular, we evaluated and compared four different types of controllers: 
cDBS, RL, RL with policy distillation (\textit{i.e.}, distilled RL), and no-DBS (i.e., without stimulation). 
The electrodes of the DBS device were placed in STN and GPi brain regions for all four participants; LFPs were sensed from STN and stimuli were delivered to both STN and GP.

Each participant also has had \textit{different PD symptoms and severity}; their characteristics are summarized in Appendix~\ref{app:patient}. All trials were conducted under close supervision of clinical experts, {strictly following the process approved by the Duke University Medical Center IRB protocol complying with the obtained FDA IDE (G180280).} 
Further, all participants provided informed written consent.

\subsection{Therapy Efficacy and Energy-Efficiency of the RL Control Policies}

We follow the offline RL and policy distillation methodology introduced in Sec.~\ref{sec:RL} to train and update (distilled) RL policies iteratively over time. Specifically, each participant 
had monthly clinical visit, where during each day of trials a total of 2-4 RL policies would be tested. A cDBS session was placed in between any two RL sessions as a control group.  
A small number of no-DBS sessions, with DBS stimulation fully off, were also tested, 
to validate our choice of the employed QoCs metrics -- \textit{i.e.}, whether they significantly change when the participants 
are not stimulated. 

After each trial day was completed, the trajectories collected from all the sessions were added to the experience replay buffer $\mathcal{E}^\mu$ unique to each participant. 
Between two consecutive visits of each participant, her $\mathcal{E}^\mu$ was used to fine-tune the top-performing policies determined from the last trial (using smaller learning rates between $[10^{-7}, 10^{-5}]$) or to train new policies from scratch (with learning rates between $[10^{-5}, 10^{-3}]$); such policies were then 
tested in the next visit. We followed~\cite{lillicrap2015continuous} and 
used two-layer NNs with 400 and 300 nodes each to parameterize the RL policies; moreover, a distilled version (student) of each corresponding full-sized RL policy (teacher) were trained 
as introduced in Sec.~\ref{subsec:distillation}, 
with each represented as a two-layer NN with 20 and 10 nodes. 
The constants in~\eqref{eq:per_step_reward} were set to $r_a=0, r_b=-1, C_1=0.3$ for all participants.

In each testing session, to evaluate the overall performance of 
the employed control policy, a total of 5 metrics were considered: the energy used by the IPG for stimulation,  
the mean beta amplitude over the session, and the 3 QoCs introduced in Sec.~\ref{subsec:setup}; for $QoC_{grasp}$, we captured the grasp frequencies of the hand that best correlates with the PD symptom for the participant (see Appendix~\ref{app:patient}~for~details).

\begin{table}[t!]
\begin{tabular}{@{}lllll@{}}
\toprule
          & cDBS & RL  & Distilled RL & No-DBS \\ \midrule
Participant 1 & 84   & 97  & 97           & 36      \\
Participant 2 & 145  & 80  & 182          & 52      \\
Participant 3 & 135  & 115 & 115          & 39      \\
Participant 4 & 124  & 119 & 98           & 48      \\ \bottomrule
\end{tabular}
\caption{Overall time, in minutes, spent toward testing each type of controller in clinical trials. Each testing session lasted 5-20 minutes, and no-DBS sessions were usually 5-min long to minimize the discomfort participants may experience.}
\vspace{-20pt}
\label{tab:trial_time}
\end{table}

Fig.~\ref{fig:overall_results} summarizes the obtained results, and Table~\ref{tab:trial_time} documents the total amount of time each controller was tested in clinic. Wilcoxon rank-sum tests~\cite{mann1947test} between cDBS and each of the other controllers were used to test the null hypothesis~--~\textit{if two sets of measurements were drawn from the same distribution} (i.e., that the controllers perform similarly over
the considered metrics); from this, 
$p$-values can be calculated.
The $p$-values accepting/rejecting the null hypothesis in the desired direction are highlighted in Fig.~\ref{fig:overall_results}. Specifically, it can be observed that, compared to cDBS, the RL policies and their distilled version can save significant (20\%-55\%) stimulation energy across participants; as $p<.05$ achieved for all participants, which rejected the null hypothesis. 

When considering the other 4 metrics, there exist a great majority of results with $p\geq.05$, accepting the null hypothesis and indicating that both RL controllers attain control (i.e., therapy) efficacy similar to cDBS. In contrast, for the no-DBS sessions, the null hypothesis is rejected in most cases. Specifically, $p<.05$ attained by no-DBS over the mean beta amplitude, for all participants, show that beta amplitudes can change significantly when sufficient DBS is received or not, which justify our choice of using the beta amplitudes to constitute MDP states. This also shows that the RL policies can follow the reward function (from Sec.~\ref{subsec:mdp}) to effectively optimize the control strategies, 
with beta amplitudes also playing an important role. Consequently, the results show that both full and distilled RL policies 
can significantly reduce the stimulation energy, 
while achieving non-inferior control efficacy compared to cDBS.

\subsubsection{Computational Complexity and Overall Energy Consumption}

We also study the additional computation time and battery consumption of the DBS system 
due the use of full-sized RL policies or their distilled version. A Surface Go 
with an Intel Pentium Gold 4415Y CPU and 4GB RAM was used as the research tablet in Fig.~\ref{fig:clinical_setup}. 
The computation time was quantified as the time needed to run a single forward pass of the NN that represents the 
RL policy. We evaluate the forward passes for both types of RL policies 200 times; 
Table~\ref{tab:comp_time} summarizes the
mean and standard deviation of the obtained computation times. 
As can be seen, the distilled RL policy can be evaluated significantly faster than its counterpart. 

Moreover, we quantify the overall battery consumption of the entire DBS system 
as the time for which the tablet or the IPG battery 
drains from 100\% to 10\% (whichever comes first). We compare the battery runtime among the full RL and distilled RL, as well as a random controller that sets the IPG to stimulate with an arbitrary amplitude in each control cycle. 
Each experiment was repeated 3 times, resulting in the statistics in Table~\ref{tab:battery} showing that the two RL-based controllers do not drastically shorten the runtime of the DBS system; ~\textit{i.e.}, the 
energy used for 
RL-based control
does not dominate the overall energy used by 
the DBS system.

\begin{table}[t!]
\begin{tabular}{@{}lll@{}}
\toprule
                           & \begin{tabular}[c]{@{}l@{}}RL Policy\\ (400$\times$300 NN)\end{tabular} & \begin{tabular}[c]{@{}l@{}}Distilled RL Policy\\ (20$\times$10 NN)\end{tabular} \\ \midrule
Mean of Computation Time & 4.78 ms                                                                 & 2.98 ms                                                                         \\
Std of Computation Time  & 32.26 ms                                                                & 1.72 ms                                                                         \\ \bottomrule
\end{tabular}
\caption{Computation time of the original RL versus the distilled RL policy.} 
\vspace{-20pt}
\label{tab:comp_time}
\end{table}

\begin{table}[t!]
\begin{tabular}{@{}llll@{}}
\toprule
             & RL          & Distilled RL & Random Controller \\ \midrule
Battery Runtime (m) & 227 $\pm$ 5 & 220 $\pm$ 6  & 247 $\pm$ 4       \\ \bottomrule
\end{tabular}
\caption{Overall battery runtime of the DBS system when the RL, distilled RL or random controllers were used. }
\vspace{-18pt}
\label{tab:battery}
\end{table}

\subsection{Evaluation of the OPE Methodology}
\label{subsec:eval_ope}

For each participant, a DLSM was trained following the methodology introduced in Sec.~\ref{sec:OPE}, 
and then 
used 
as a synthetic environment 
to interact with 6 policies trained using the deep actor-critic method (Sec.~\ref{sec:RL}) with different hyper-parameters, over the buffer $\mathcal{E}^\mu$ specific to the patient; these policies can in general lead to varying performance. Then, for each policy, the mean of total returns~\eqref{eq:ope_total_return} over all simulated trajectories can be calculated, and was used to estimate the policy's expected return from Problem~\ref{prob:ope_objective}.
The constants in~\eqref{eq:end_of_session_reward}, balancing the scale of the QoCs (\textit{i.e.}, grasp frequency, rating and tremor severity) were set to $C_2 = C_3 = C_4 = 10$ for patients~2-4 who can experience bradykinesia and pronounced tremor~with~insufficient DBS; in contrast, the symptoms of participant 1 are considered subtle, so we set $C_2 = C_3 = C_4 = 25$ to better distinguish if sufficient DBS is provided; see Appendix~\ref{app:patient} for details on patient characteristics {as well as the dosage of PD medications}.

DLSM's performance was compared against the classic IS~\cite{precup2000eligibility}, as well as a state-of-the-art IS-based OPE method, dual-DICE~\cite{nachum2019dualdice}. Three metrics were considered to evaluate the performance of OPE, including mean absolute error (MAE), rank correlation, and regret@1, following from~\cite{fu2020benchmarks}.
MAE evaluates the absolute error between the total return estimated by OPE, versus the \textit{actual} returns,~\textit{i.e.}, mean total return recorded from clinical trials. Rank correlation quantifies the alignment between the rank of policies over OPE-estimated returns and the actual returns. Regret@1 quantifies the percentage loss, over the total actual returns, one would get by picking the policy with maximum OPE-estimated return, against the actual best-performing policy, showing if the OPE methods can identify the best-performing policy correctly. Their mathematical definitions can be found in Appendix~\ref{app:ope_metrics}. 

\begin{figure}[t!]
    \centering
    \includegraphics[width=.94\linewidth]{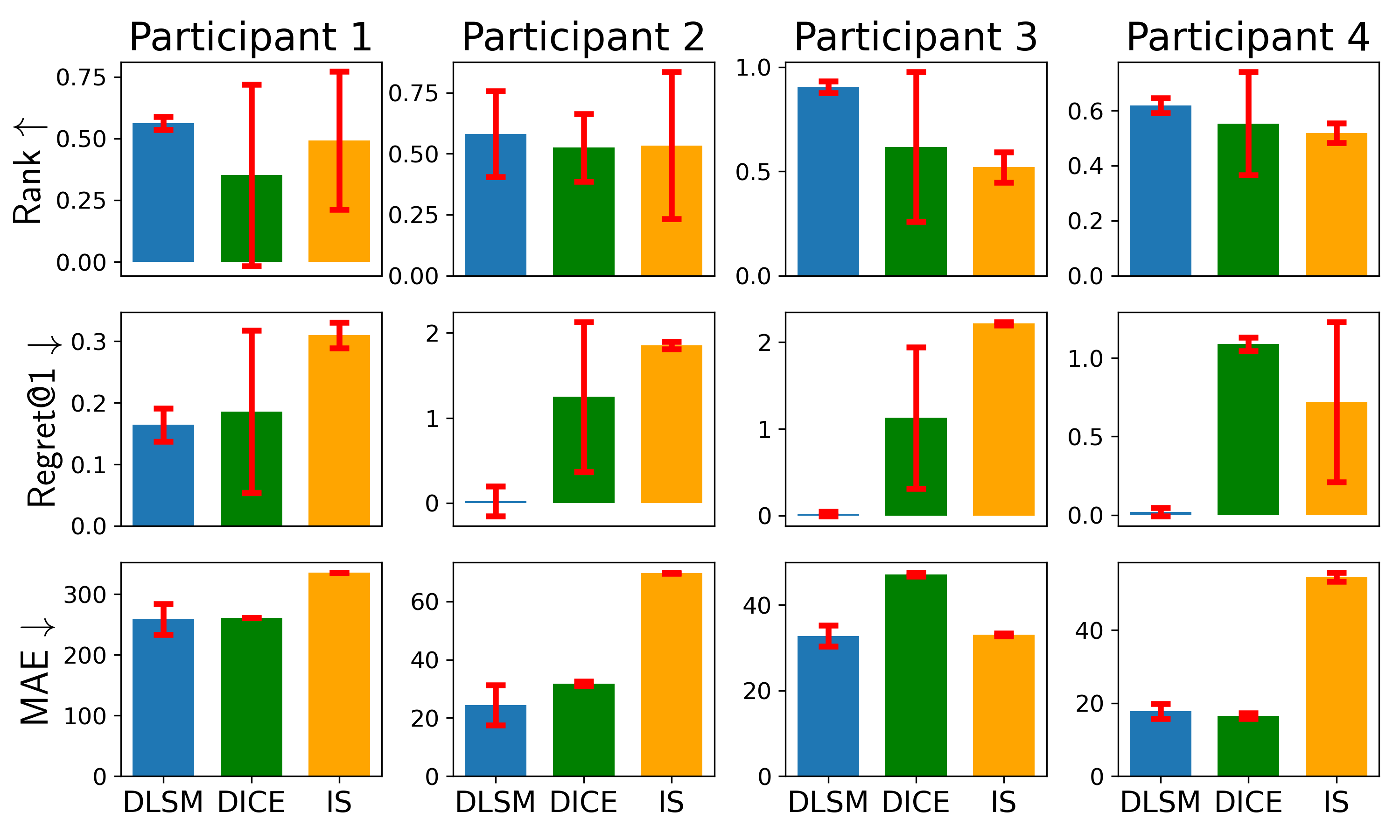}
    \vspace{-10pt}
    \caption{DLSM in general achieves higher ranks, lower regret@1's and lower MAEs, compared to DICE and IS. Each method is trained and evaluated with 3 different random seeds, with the standard deviations shown by the error bars.}
    \vspace{-4pt}
    \label{fig:ope_results}
\end{figure}

The obtained results are summarized in Fig.~\ref{fig:ope_results}. As shown, the DLSM in general achieved significantly higher rank and lower regret, as well as non-inferior MAE, over DICE and IS.

%% file: Discussion.tex
\section{Conclusion}
\label{sec:discussion}

In this paper, we introduced an offline RL and OPE framework 
to design and evaluate closed-loop DBS controllers using only historical data. 
Moreover, a policy distillation method was introduced to further reduce the computation requirements for evaluating RL policies. The control efficacy and energy efficiency of the RL controllers were validated with clinical testing over 4 patients. Results showed that 
RL-based controllers lead to similar 
control efficacy as cDBS, but with significantly reduced stimulation energy. The computation times for 
the RL and distilled RL controllers were compared, showing that the distilled version executed significantly faster; future work will focus on further reducing execution times of the distilled RL controllers to match capabilities of implanted devices. 
Finally, the DLSM is trained to estimate the expected returns of RL policies, which outperforms existing IS-based OPE methods, in terms of rank correlations, regrets~and~MAEs.